\documentclass[journal]{IEEEtran}
\usepackage[T1]{fontenc}
\usepackage[utf8]{inputenc}
\usepackage{amsmath,amssymb}
\usepackage{graphicx}
\usepackage{booktabs}
\usepackage{hyperref}
\usepackage{url}
\usepackage{microtype}
\usepackage{xcolor}
\usepackage{siunitx}
\usepackage{multirow}
\usepackage{caption}

\hypersetup{
  colorlinks=true,
  linkcolor=blue!60!black,
  citecolor=blue!60!black,
  urlcolor=blue!60!black
}

\begin{document}

\title{Do Foundation Model Embeddings Improve Cross-Country Crop Yield Generalisation?
       A Leave-One-Country-Out Evaluation in Sub-Saharan Africa}

\author{Yaw~Osei~Adjei\\
  Department of Computer Science,\\
  Kwame Nkrumah University of Science and Technology,\\
  Kumasi, Ghana\\
  \texttt{yoadjei@st.knust.edu.gh}
}

\maketitle

\begin{abstract}
Accurate smallholder maize yield prediction across national boundaries is critical
for food security planning in sub-Saharan Africa, yet most published benchmarks
report within-country performance that overstates true generalisability.
This paper asks whether geospatial foundation model embeddings —
specifically Prithvi-EO-1.0-100M and ViT-Base — outperform traditional
Sentinel-2 spectral features under a rigorous Leave-One-Country-Out (LOCO) cross-validation
scheme applied to 6{,}404 smallholder maize field observations across
five African countries (Kenya, Malawi, Nigeria, Rwanda, Tanzania) for
the years 2017--2022.
We evaluate 18 experimental conditions spanning three feature representations,
three regression algorithms (Ridge, Random Forest, XGBoost), and two
validation protocols (LOCO and standard five-fold random cross-validation).
Results reveal a stark generalisability gap:
all feature sets achieve R$^2 \in [0.17, 0.30]$ under within-country evaluation
but universally negative R$^2 \in [-0.09, -0.03]$ under LOCO,
including a naive country-mean baseline.
Critically, frozen Prithvi-EO embeddings applied to single growing-season
composites offer no meaningful advantage over 10-band Sentinel-2 band medians
for cross-country prediction.
This result must be interpreted carefully: Prithvi-EO was designed for
multi-temporal input, and using a single annual composite is a deliberate
mismatch that isolates the question of whether frozen representations alone
carry cross-country invariance.
We find they do not, and interpret this as evidence that the primary bottleneck
is country-level yield distribution shift — a data problem — rather than
deficiencies in feature representation.
Our findings challenge the assumption that domain-specific geospatial
pre-training closes the generalisability gap, and establish a reproducible
negative benchmark that future work should surpass.
All code and processed results are released at
\url{https://github.com/yoadjei/yield-africa}.
\end{abstract}


\section{Introduction}
\label{sec:intro}

Food insecurity remains a defining challenge across sub-Saharan Africa,
where smallholder farms account for more than 70\% of staple crop production~\cite{burke2017sat}.
Timely and spatially disaggregated yield forecasts are essential inputs for
early warning systems, agricultural policy, and targeted humanitarian response.
Remote sensing offers a scalable, cost-effective route to such forecasts by
linking satellite observations directly to field-level yield outcomes~\cite{lobell2020eyes}.

A central ambition of this programme is to train predictive models in
data-rich countries and apply them in data-scarce settings — a transfer
that requires genuine cross-country generalisation.
Yet the dominant evaluation paradigm in the literature applies random
$k$-fold cross-validation within a pooled dataset, inadvertently preserving
country-specific statistical patterns in both training and test folds.
This produces optimistic accuracy estimates that collapse when
models are deployed across national boundaries.

The recent emergence of geospatial foundation models pre-trained on
large-scale satellite archives has raised hopes that self-supervised
representations might encode invariant physical features of agricultural
landscapes, thereby reducing this generalisation gap.
Prithvi-EO-1.0-100M~\cite{jakubik2023prithvi}, jointly released by IBM and NASA,
is a Masked Autoencoder~\cite{he2022mae} pre-trained on
six-channel Harmonised Landsat-Sentinel (HLS) time series specifically
for crop and land-cover applications.
The Vision Transformer (ViT-Base)~\cite{dosovitskiy2021vit}, pre-trained
on ImageNet at scale, provides a general-purpose visual baseline for comparison.

This paper makes three contributions:

\begin{enumerate}
  \item We present \textbf{a systematic LOCO evaluation} of frozen foundation
        model embeddings against engineered spectral features for smallholder
        maize yield prediction in Africa, using 6{,}404 observations from
        five countries spanning 2017--2022.
        To our knowledge, no prior published work applies this evaluation
        protocol to Prithvi-EO or comparable geospatial foundation models
        in the African smallholder yield regression setting.
  \item We demonstrate that \textbf{all feature representations fail to generalise
        cross-country under the frozen, single-frame evaluation protocol}:
        LOCO R$^2$ is universally negative for all 9 combinations
        of feature set and regressor, and a naive country-mean predictor also
        yields negative R$^2$, confirming the primary bottleneck is distribution
        shift in yield itself rather than in the feature space.
  \item We provide a \textbf{reproducible negative benchmark} — a clearly documented
        failure mode that future work on domain adaptation,
        meta-learning, and cross-country yield transfer must surpass.
\end{enumerate}

The remainder of this paper is organised as follows.
Section~\ref{sec:related} reviews related work.
Section~\ref{sec:data} describes the study area and data sources.
Section~\ref{sec:methods} presents the experimental design.
Section~\ref{sec:results} reports results.
Section~\ref{sec:discussion} discusses implications, and
Section~\ref{sec:conclusion} concludes.

\section{Related Work}
\label{sec:related}

\subsection{Satellite-Based Yield Prediction in Africa}

Burke and Lobell~\cite{burke2017sat} demonstrated that Landsat-derived NDVI time
series can explain a significant fraction of smallholder yield variation in
Uganda, establishing early proof of concept for satellite-based prediction in Africa.
Lobell et al.~\cite{lobell2020eyes} conducted a systematic comparison of
satellite imagery and traditional household surveys, finding that
imagery-based prediction achieves moderate accuracy within-country but
highlighted the challenge of transferring models across contexts.
You et al.~\cite{you2017deep} applied deep Gaussian processes to MODIS time
series for county-level yield prediction in the United States, showing that
spatial dependencies are a key driver of predictive accuracy.
These studies share a common limitation: cross-location evaluation is rare,
and cross-country evaluation is nearly absent.

\subsection{Foundation Models for Geospatial Applications}

The Vision Transformer~\cite{dosovitskiy2021vit}, through its attention mechanism,
demonstrated that large-scale pre-training on natural images enables powerful
general-purpose visual representations.
Mai et al.~\cite{mai2023foundation} surveyed opportunities and challenges of
foundation models for geospatial AI, noting a persistent gap between
general pre-training objectives and specialised downstream tasks.
Jakubik et al.~\cite{jakubik2023prithvi} introduced Prithvi-EO-1.0-100M,
a 100-million parameter masked autoencoder pre-trained on six-channel HLS imagery
across the continental United States and fine-tuned for flood mapping,
multi-temporal crop segmentation, and burned area detection.
Wang et al.~\cite{wang2022empirical} curated the SSL4EO-S12 benchmark for
self-supervised pre-training on Sentinel-1/2 imagery, highlighting the
importance of modality alignment between pre-training and downstream tasks.
No prior work has benchmarked frozen Prithvi-EO embeddings against
spectral baselines for \emph{cross-country} yield regression in Africa.

\subsection{Domain Shift and Cross-Location Transfer}

Tseng et al.~\cite{tseng2021crop} introduced CropHarvest, a global crop-type
dataset, and evaluated few-shot cross-country crop classification,
finding severe performance degradation without target-country adaptation.
Kerner et al.~\cite{kerner2020rapid} developed rapid-response crop mapping
for data-sparse regions, but relied on within-region fine-tuning.
Tuia et al.~\cite{tuia2016domain} provided a comprehensive survey of domain
adaptation in remote sensing classification, establishing that covariate shift
between acquisition conditions routinely degrades cross-site generalisation;
their findings extend conceptually to the regression setting studied here.
Wolanin et al.~\cite{wolanin2020estimating} demonstrated that explainable
deep learning can achieve strong within-country yield estimates in the Indian
wheat belt, yet their evaluation was confined to a single country and did
not address cross-country transfer.
Ma\~{n}as et al.~\cite{manas2021seasonal} showed that seasonal contrast
pre-training on uncurated RS data improves linear probe accuracy across
multiple downstream RS tasks, but did not evaluate cross-country yield regression.
These studies reinforce a consistent pattern: geospatial models trained in
one country or region degrade substantially when applied elsewhere without
adaptation.
Our work extends this evidence to the yield regression setting and explicitly
quantifies the gap using a LOCO protocol across five African countries,
adding the dimension of comparing foundation model embeddings against
engineered spectral features under the same LOCO protocol.

\section{Study Area and Data}
\label{sec:data}

\subsection{Countries and Crop}

The study covers five sub-Saharan African countries:
Kenya, Malawi, Nigeria, Rwanda, and Tanzania.
Maize (\textit{Zea mays}) is the target crop, selected as the most widely
cultivated staple in the region and the crop with the highest GROW-Africa
observation density.
The five countries represent diverse agroecological zones
(semi-arid savanna to highland tropics) and a wide range of smallholder
yield outcomes, making cross-country generalisation genuinely challenging.

\subsection{Yield Labels}

Field-level yield labels are drawn from two complementary sources:

\textbf{GROW-Africa}~\cite{grow_africa} provides GPS-tagged, farm-level maize
yield observations collected by household surveys and agronomic trials
across sub-Saharan Africa.
We applied the following filters: (i) observation years 2017--2022 to align
with Sentinel-2 availability; (ii) point-level GPS coordinates only, excluding
administrative-polygon centroids; (iii) countries with at least 100
post-filter observations.

\textbf{HarvestStat Africa}~\cite{lee2025harveststat} provides subnational
crop production statistics.
Nigeria has negligible GPS-level GROW-Africa coverage; HarvestStat admin-unit
centroids were used as proxy observations for that country.
A \texttt{label\_source} indicator distinguishes the two data types.
This coarser spatial resolution for Nigeria is a disclosed limitation
(see Section~\ref{sec:limits}).

After merging and quality filtering, the dataset contains
\textbf{6{,}404 observations}:
Kenya (1{,}396), Malawi (1{,}552), Nigeria (955),
Rwanda (1{,}138), and Tanzania (1{,}363).
Table~\ref{tab:dataset} summarises the per-country distribution.

\begin{table}[!t]
  \centering
  \caption{Per-country dataset summary. Yield in kg/ha on original scale.}
  \label{tab:dataset}
  \begin{tabular}{lrrrrr}
    \toprule
    Country   & $n$ & Mean & SD   & Min  & Max    \\
    \midrule
    Kenya     & 1{,}396 & 3{,}278 & 1{,}592 & 132  & 8{,}278  \\
    Malawi    & 1{,}552 & 1{,}918 & 1{,}366 & 123  & 6{,}919  \\
    Nigeria   & 955     & 1{,}256 & 1{,}222 & 122  & 6{,}203  \\
    Rwanda    & 1{,}138 & 3{,}993 & 1{,}874 & 360  & 9{,}777  \\
    Tanzania  & 1{,}363 & 3{,}255 & 1{,}669 & 173  & 8{,}500  \\
    \midrule
    Total     & 6{,}404 & 2{,}769 & 1{,}825 & 122  & 9{,}777  \\
    \bottomrule
  \end{tabular}
\end{table}

The wide inter-country yield variation (Nigeria mean 1{,}256 kg/ha vs.\
Rwanda mean 3{,}993 kg/ha) is a key driver of the distribution shift studied here.

\subsection{Sentinel-2 Imagery}

Sentinel-2 Level-2A surface reflectance imagery was acquired via Google Earth Engine
for a 500~m buffer around each field centroid, composited as annual growing-season
medians~\cite{drusch2012sentinel2}.
Ten spectral bands were extracted: B2 (blue), B3 (green), B4 (red),
B5--B7 (red-edge), B8 (NIR), B8A (narrow NIR), B11 and B12 (SWIR).
A 224~$\times$~224 pixel patch centred on each field was exported as a
GeoTIFF for foundation model inference.
Patches with estimated cloud cover exceeding 20\% were discarded.

\subsection{Spectral Indices}

Four spectral indices were computed from the Sentinel-2 composites:
NDVI~\cite{rouse1974ndvi} (Normalised Difference Vegetation Index),
EVI~\cite{jiang2008evi} (Enhanced Vegetation Index),
LSWI (Land Surface Water Index), and NDWI (Normalised Difference Water Index).
These indices capture vegetation greenness and canopy moisture status
relevant to yield formation.

\subsection{Rainfall}

Growing-season cumulative precipitation was extracted from the CHIRPS~\cite{funk2015chirps}
daily dataset at 0.05° resolution. Three features were derived per field:
total seasonal rainfall, mean daily rainfall, and the coefficient of variation
(a proxy for drought stress).

\section{Methods}
\label{sec:methods}

\subsection{Feature Representations}

Three feature sets were evaluated, each representing a distinct
data representation paradigm:

\subsubsection{Spectral (baseline)}
A hand-engineered 23-dimensional vector comprising the 10 Sentinel-2 band medians,
four spectral indices, and three CHIRPS rainfall statistics.
This represents the standard engineered-feature approach used in most
operational yield prediction systems.

\subsubsection{Prithvi-EO Embeddings}
Prithvi-EO-1.0-100M~\cite{jakubik2023prithvi} is a Vision Transformer with
Masked Autoencoder pre-training on six-channel (HLS blue, green, red, NIR,
SWIR-1, SWIR-2) multi-temporal imagery.
Sentinel-2 patches were resampled to the six HLS channels, normalised using the
model's published per-channel statistics, and fed to the frozen encoder as
a single-frame tensor of shape $(B, 6, 1, 224, 224)$.
The 768-dimensional CLS token from the final encoder block was used as the embedding.
No fine-tuning was performed; weights were loaded directly from the
publicly released checkpoint (\texttt{Prithvi\_EO\_V1\_100M.pt}).
This isolates the contribution of pre-trained representations from
task-specific adaptation.

\subsubsection{ViT-Base Embeddings}
ViT-Base/16~\cite{dosovitskiy2021vit} pre-trained on ImageNet-21k was applied to
RGB (B4, B3, B2) Sentinel-2 patches, producing a 768-dimensional CLS token.
ViT-Base serves as a general-purpose vision baseline, enabling comparison between
geospatial-specialised pre-training (Prithvi-EO) and domain-agnostic pre-training
on natural images.

\subsection{Regressors}

Three regression algorithms were evaluated:

\textbf{Ridge Regression}~\cite{hoerl1970ridge} with $\alpha \in \{0.1, 1, 10, 100, 1000\}$
selected by inner cross-validation. Ridge provides a linear baseline that is
robust to multicollinearity — relevant given the high dimensionality of
embedding features.

\textbf{Random Forest}~\cite{breiman2001rf} with 100 trees and all default
hyperparameters. Random Forest captures non-linear interactions without
being prone to overfitting individual training points.

\textbf{XGBoost}~\cite{chen2016xgboost} with 300 trees, learning rate 0.05,
maximum depth 4, and subsampling 0.8.
XGBoost is the strongest tree ensemble baseline in tabular regression benchmarks
and provides a competitive upper bound on performance within this feature paradigm.

All models were wrapped in a \texttt{StandardScaler} $\rightarrow$ regressor
pipeline. Random state was fixed at 42 across all experiments.

\subsection{Target Variable}

The yield target was log-transformed (\texttt{yield\_log}) after confirming
right skewness greater than 1.0 in the pooled distribution.
Log transformation stabilises variance and is standard practice for
smallholder yield regression. All reported RMSE values are back-transformed
to kg/ha for interpretability.

\subsection{Missing Value Imputation}

NaN values in feature vectors — arising from cloud-affected patches or
partial CHIRPS coverage — were imputed with the training-set column median.
All-NaN columns (which can occur for held-out countries with zero valid
spectral observations) were filled with 0. Rows with NaN yield values were dropped.

\subsection{Cross-Validation Schemes}

Two orthogonal evaluation protocols were applied to every feature--regressor
combination, yielding 18 experimental conditions in total.

\subsubsection{Random Five-Fold CV}
Standard stratified five-fold cross-validation with shuffling.
This protocol leaks country-specific signal into the training folds
and measures within-distribution predictive accuracy.

\subsubsection{Leave-One-Country-Out (LOCO) CV}
For each of the five countries, the model is trained on the remaining
four countries and evaluated on the held-out country.
Per-country RMSE, MAE, and R$^2$ are recorded, and predictions are
concatenated across folds to produce aggregate metrics.
LOCO is the appropriate evaluation protocol for the stated goal of
cross-country generalisation.

\subsection{Naive Baseline}

A non-parametric baseline was constructed for each held-out country
by predicting all test observations with the mean yield of the
corresponding training set (all other countries).
Because country yield means differ substantially, this baseline also
achieves negative R$^2$, confirming that the challenge is not merely
underfitting but genuine distribution shift.

\subsection{Evaluation Metrics}

Three metrics are reported:
root mean squared error (RMSE, kg/ha),
mean absolute error (MAE, kg/ha), and
the coefficient of determination R$^2$.
R$^2 < 0$ indicates the model performs worse than predicting the
test-set mean yield — a more severe failure than simply underfitting.

\section{Results}
\label{sec:results}

\subsection{Within-Country Performance (Random CV)}

\begin{figure}[!t]
  \centering
  \includegraphics[width=\columnwidth]{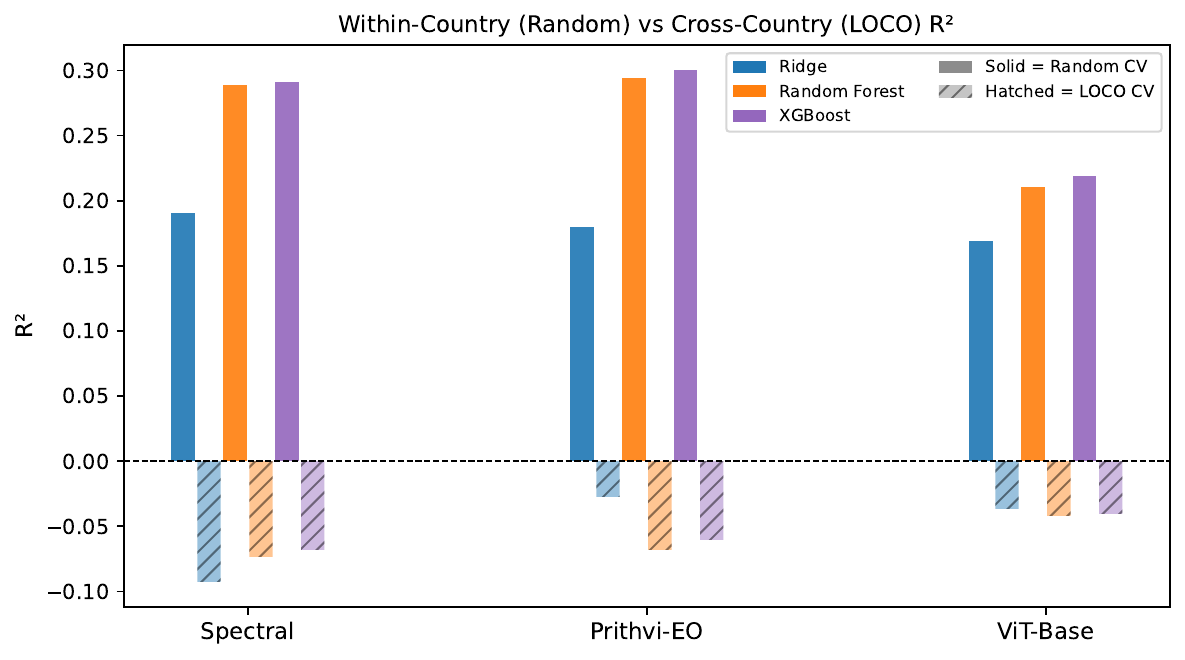}
  \caption{Within-country (random CV) vs.\ cross-country (LOCO) R$^2$ for all
           nine feature--regressor combinations. Every condition shows a large
           drop when moving from random to LOCO evaluation.}
  \label{fig:random_vs_loco}
\end{figure}

Table~\ref{tab:random} and Figure~\ref{fig:random_vs_loco} report five-fold random CV results.
All nine feature--model combinations achieve positive R$^2$,
with values ranging from 0.169 (ViT-Base / Ridge) to 0.300
(Prithvi-EO / XGBoost).
Tree ensemble methods (RF and XGBoost) consistently outperform Ridge
regardless of feature set, consistent with non-linear yield--feature
relationships.
Prithvi-EO and spectral features are nearly equivalent:
best-case Prithvi-EO R$^2 = 0.300$ vs.\ spectral R$^2 = 0.291$,
a difference of 0.009 R$^2$ units.
ViT-Base lags behind both domain-specific representations (best R$^2 = 0.219$),
suggesting that ImageNet pre-training provides less useful inductive
biases for vegetation analysis than geospatial pre-training.

\begin{table}[!t]
  \centering
  \caption{Random five-fold CV results. Best R$^2$ per feature set in \textbf{bold}.}
  \label{tab:random}
  \setlength{\tabcolsep}{5pt}
  \begin{tabular}{llrrr}
    \toprule
    Feature Set  & Regressor & RMSE (kg/ha) & MAE (kg/ha) & R$^2$  \\
    \midrule
    Spectral     & Ridge     & 1641.9       & 1308.7      & 0.191  \\
    Spectral     & RF        & 1539.1       & 1185.9      & 0.289  \\
    Spectral     & XGBoost   & 1536.7       & 1211.1      & \textbf{0.291}  \\
    \midrule
    Prithvi-EO   & Ridge     & 1652.7       & 1318.2      & 0.180  \\
    Prithvi-EO   & RF        & 1533.4       & 1188.1      & 0.294  \\
    Prithvi-EO   & XGBoost   & 1527.0       & 1197.3      & \textbf{0.300}  \\
    \midrule
    ViT-Base     & Ridge     & 1663.8       & 1325.4      & 0.169  \\
    ViT-Base     & RF        & 1621.3       & 1273.2      & 0.211  \\
    ViT-Base     & XGBoost   & 1612.8       & 1273.0      & \textbf{0.219}  \\
    \bottomrule
  \end{tabular}
\end{table}

\subsection{Cross-Country Generalisation (LOCO CV)}

\begin{figure}[!t]
  \centering
  \includegraphics[width=\columnwidth]{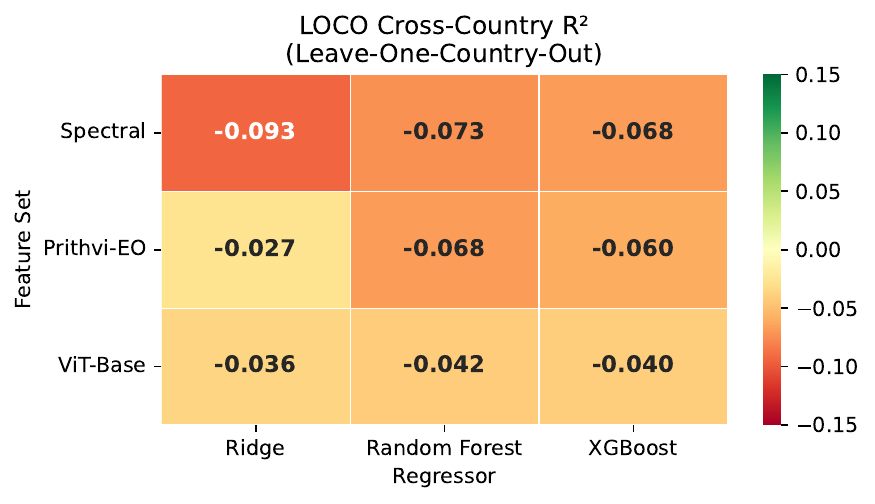}
  \caption{LOCO R$^2$ heatmap across all nine feature--regressor combinations.
           All values are negative. Prithvi-EO / Ridge achieves the
           least-negative result ($-0.027$).}
  \label{fig:loco_heatmap}
\end{figure}

Table~\ref{tab:loco} and Figure~\ref{fig:loco_heatmap} report LOCO results.
All 9 conditions yield negative R$^2$, ranging from $-0.093$
(Spectral / Ridge) to $-0.027$ (Prithvi-EO / Ridge).
The aggregate RMSE (1{,}850--1{,}910 kg/ha) is notably higher than
the random CV RMSE (1{,}527--1{,}664 kg/ha), reflecting the additional
difficulty of predicting across country boundaries.

\begin{table}[!t]
  \centering
  \caption{LOCO CV results. Least-negative R$^2$ per feature set in \textbf{bold}.}
  \label{tab:loco}
  \setlength{\tabcolsep}{5pt}
  \begin{tabular}{llrrr}
    \toprule
    Feature Set  & Regressor & RMSE (kg/ha) & MAE (kg/ha) & R$^2$   \\
    \midrule
    Spectral     & Ridge     & 1907.8       & 1498.2      & $-0.093$ \\
    Spectral     & RF        & 1890.7       & 1547.6      & $-0.073$ \\
    Spectral     & XGBoost   & 1885.9       & 1536.8      & \textbf{$-0.068$} \\
    \midrule
    Prithvi-EO   & Ridge     & 1849.5       & 1494.0      & \textbf{$-0.027$} \\
    Prithvi-EO   & RF        & 1885.8       & 1551.6      & $-0.068$ \\
    Prithvi-EO   & XGBoost   & 1879.4       & 1532.0      & $-0.060$ \\
    \midrule
    ViT-Base     & Ridge     & 1858.0       & 1508.2      & \textbf{$-0.036$} \\
    ViT-Base     & RF        & 1863.1       & 1536.2      & $-0.042$ \\
    ViT-Base     & XGBoost   & 1861.7       & 1524.8      & $-0.040$ \\
    \bottomrule
  \end{tabular}
\end{table}

\subsection{Naive Baseline}

\begin{figure}[!t]
  \centering
  \includegraphics[width=\columnwidth]{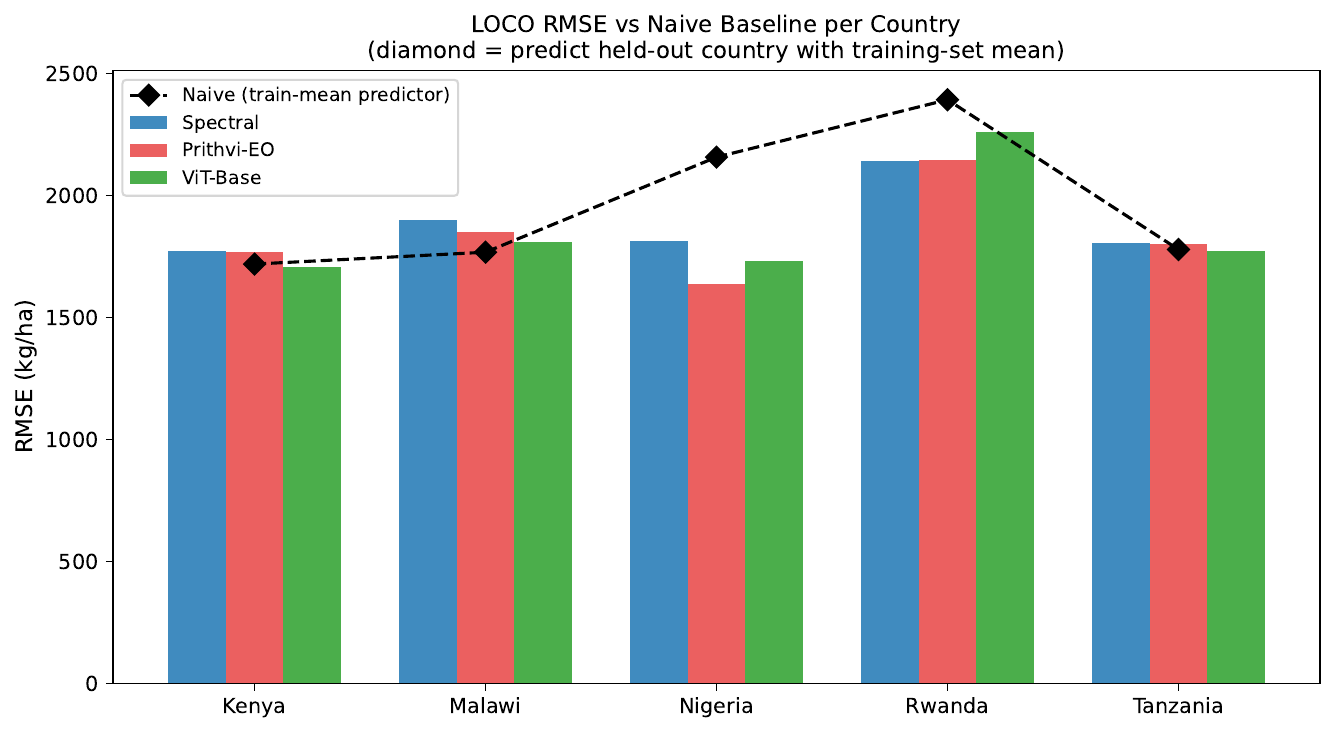}
  \caption{Per-country RMSE for the best model per feature set (bars) vs.\
           the naive country-mean baseline (diamond, dashed line).
           Learned models beat the naive baseline on RMSE, but all remain
           below zero R$^2$.}
  \label{fig:naive}
\end{figure}

Table~\ref{tab:naive} confirms that the naive country-mean predictor
also achieves universally negative R$^2$ under LOCO.
The naive RMSE (1{,}719--2{,}392 kg/ha) exceeds the best-model RMSE
for all countries, indicating that learned representations do provide
some signal beyond the simplest possible predictor.
However, both learned models and the naive baseline lie below zero R$^2$,
confirming that predicting held-out countries from other-country data
is a fundamentally difficult task regardless of model sophistication.

\begin{table}[!t]
  \centering
  \caption{Naive country-mean baseline under LOCO.}
  \label{tab:naive}
  \begin{tabular}{lrrrr}
    \toprule
    Country   & $n_{\text{test}}$ & Naive RMSE (kg/ha) & Naive R$^2$ \\
    \midrule
    Kenya     & 1{,}396           & 1{,}719            & $-0.167$   \\
    Malawi    & 1{,}552           & 1{,}768            & $-0.676$   \\
    Nigeria   & 955               & 2{,}157            & $-2.121$   \\
    Rwanda    & 1{,}138           & 2{,}392            & $-0.631$   \\
    Tanzania  & 1{,}363           & 1{,}779            & $-0.137$   \\
    \bottomrule
  \end{tabular}
\end{table}

\subsection{Generalisation Gap}

\begin{figure}[!t]
  \centering
  \includegraphics[width=\columnwidth]{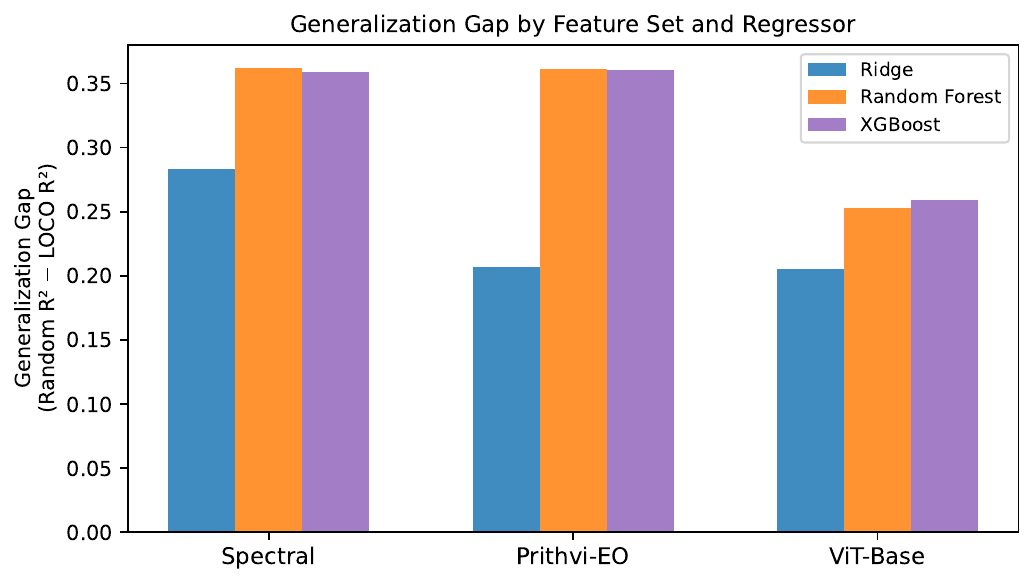}
  \caption{Generalisation gap (random CV R$^2$ minus LOCO R$^2$) per
           feature--regressor combination. Ridge consistently shows smaller
           gaps than tree ensembles despite lower within-country accuracy.}
  \label{fig:gap}
\end{figure}

The generalisation gap — defined as random R$^2$ minus LOCO R$^2$ —
is visualised in Figure~\ref{fig:gap} and ranges from 0.216 (Prithvi-EO / Ridge: $0.180 - (-0.027)$) to 0.384 (Spectral / Ridge: $0.191 - (-0.093)$).
XGBoost, despite being the strongest within-country model, exhibits
generalisation gaps of 0.359 (spectral), 0.360 (Prithvi-EO),
and 0.259 (ViT-Base), demonstrating that higher within-country accuracy
does not imply better cross-country transfer.
Ridge consistently exhibits smaller generalisation gaps than tree
ensembles, suggesting that simpler models are more robust to
distribution shift.

\subsection{NDVI-Only Ablation}

To bound the lower end of feature richness, Table~\ref{tab:ndvi}
evaluates a single-feature baseline using only NDVI under LOCO.
NDVI-only Ridge achieves R$^2 = -0.170$, worse than the 23-feature
spectral baseline ($-0.093$), confirming that the full engineered
feature set provides incremental signal even when all models fail
to generalise cross-country.
The gap between NDVI-only and the best full-feature model
(Prithvi-EO / Ridge, $-0.027$) is only 0.143 R$^2$ units —
small relative to the within-country to LOCO gap of $>0.2$ units,
which underscores that feature richness is not the binding constraint.

\begin{table}[!t]
  \centering
  \caption{NDVI-only ablation under LOCO (single feature, all three regressors).}
  \label{tab:ndvi}
  \begin{tabular}{lrrr}
    \toprule
    Regressor & RMSE (kg/ha) & MAE (kg/ha) & R$^2$ \\
    \midrule
    Ridge     & 1{,}938.3    & 1{,}539.5   & $-0.170$ \\
    RF        & 2{,}086.8    & 1{,}658.0   & $-0.532$ \\
    XGBoost   & 1{,}954.1    & 1{,}553.7   & $-0.188$ \\
    \bottomrule
  \end{tabular}
\end{table}

\subsection{Nigeria Sensitivity Analysis}

Nigeria labels are admin-centroid proxies rather than GPS-tagged field
observations (Section~\ref{sec:data}).
Table~\ref{tab:nng} reports LOCO results on the four remaining countries
after excluding Nigeria from both training and test folds.
For tree ensembles, aggregate R$^2$ remains strongly negative
(spectral/RF: $-0.156$; spectral/XGBoost: $-0.139$) while Ridge
deteriorates ($-0.278$).
This non-monotone pattern arises because Nigeria occupies a distinct
low-yield region of the training distribution that anchors the linear
decision boundary; its removal shifts the pooled training mean toward
higher yields, increasing prediction error for Malawi (mean 1{,}918 kg/ha).
Critically, all conditions remain well below zero R$^2$, confirming that
distribution shift is pervasive and not an artefact of Nigeria's
label quality.

\begin{table}[!t]
  \centering
  \caption{Nigeria-excluded LOCO sensitivity (spectral features;
           Kenya, Malawi, Rwanda, Tanzania only).}
  \label{tab:nng}
  \begin{tabular}{lrrr}
    \toprule
    Regressor & RMSE (kg/ha) & MAE (kg/ha) & R$^2$ \\
    \midrule
    Ridge     & 2{,}033.8    & 1{,}650.4   & $-0.278$ \\
    RF        & 1{,}920.2    & 1{,}578.3   & $-0.156$ \\
    XGBoost   & 1{,}907.5    & 1{,}567.1   & $-0.139$ \\
    \bottomrule
  \end{tabular}
\end{table}

\subsection{Per-Fold Variability}

With only five LOCO folds, aggregate R$^2$ figures mask large
per-country variance.
Table~\ref{tab:foldstd} reports the mean and standard deviation of
per-country R$^2$ across the five country holdouts.
Standard deviations of 0.32--0.70 R$^2$ units dwarf the differences
between feature sets ($<0.07$ aggregate R$^2$ units) and between
regressors ($<0.05$ units).
Differences between conditions in Tables~\ref{tab:random}
and~\ref{tab:loco} should therefore be interpreted as indicative
directional trends, not statistically separable findings.

\begin{table}[!t]
  \centering
  \caption{Per-country R$^2$ mean $\pm$ std across the five LOCO holdouts.
           Large std values confirm that aggregate results are driven by
           which country is hardest to predict, not by small feature differences.}
  \label{tab:foldstd}
  \setlength{\tabcolsep}{4pt}
  \begin{tabular}{llrr}
    \toprule
    Feature Set & Regressor & Mean R$^2$ & Std R$^2$ \\
    \midrule
    Spectral    & Ridge     & $-0.548$ & $\pm 0.542$ \\
    Spectral    & RF        & $-0.603$ & $\pm 0.568$ \\
    Spectral    & XGBoost   & $-0.572$ & $\pm 0.468$ \\
    \midrule
    Prithvi-EO  & Ridge     & $-0.470$ & $\pm 0.322$ \\
    Prithvi-EO  & RF        & $-0.634$ & $\pm 0.696$ \\
    Prithvi-EO  & XGBoost   & $-0.594$ & $\pm 0.587$ \\
    \midrule
    ViT-Base    & Ridge     & $-0.500$ & $\pm 0.384$ \\
    ViT-Base    & RF        & $-0.546$ & $\pm 0.545$ \\
    ViT-Base    & XGBoost   & $-0.516$ & $\pm 0.433$ \\
    \bottomrule
  \end{tabular}
\end{table}

\subsection{Per-Country Analysis}

\begin{figure}[!t]
  \centering
  \includegraphics[width=\columnwidth]{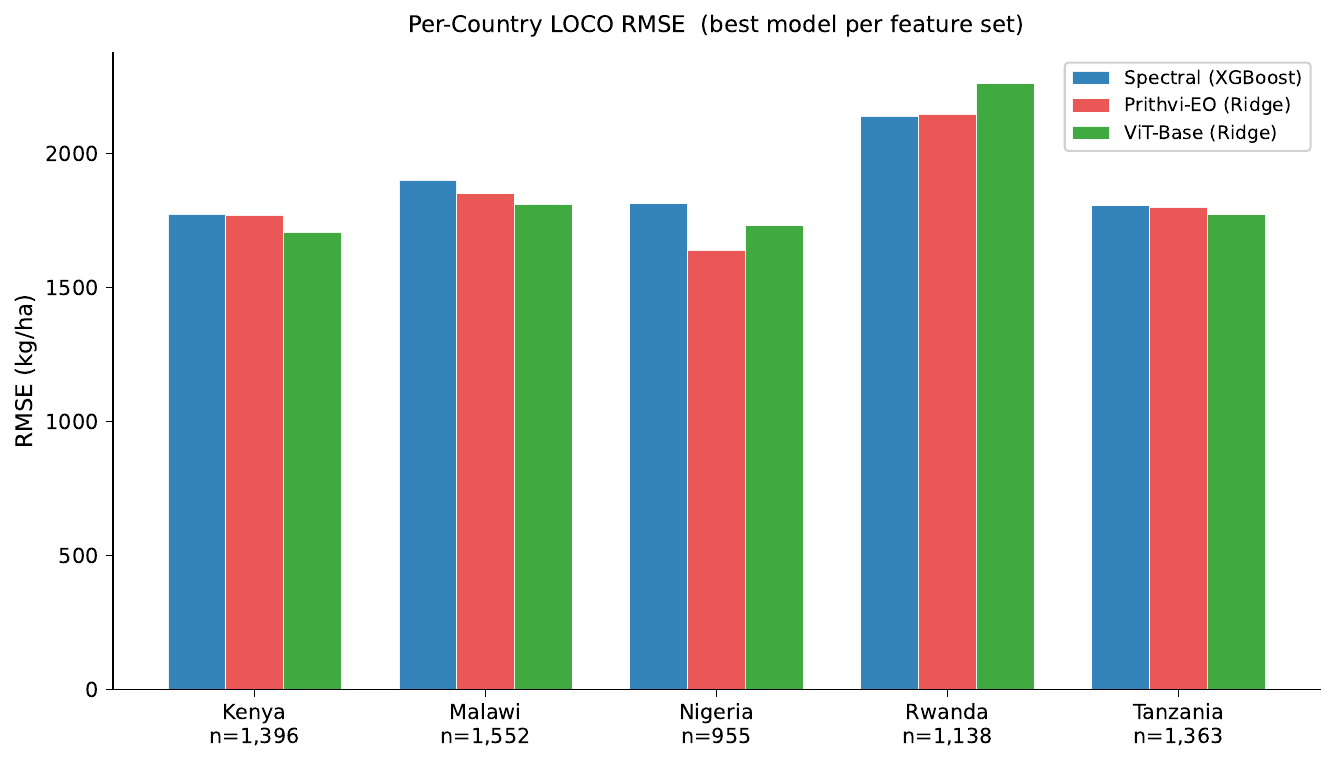}
  \caption{Per-country LOCO RMSE (kg/ha) for all nine conditions, with
           test-set sample sizes annotated. Rwanda and Nigeria are
           consistently the most difficult held-out countries.}
  \label{fig:country_rmse}
\end{figure}

\begin{figure*}[!t]
  \centering
  \includegraphics[width=\textwidth]{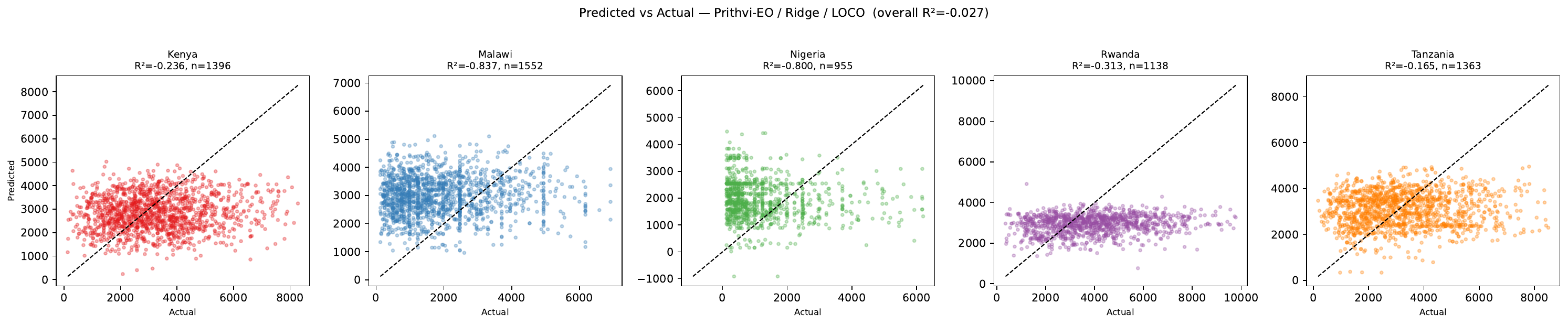}
  \caption{Predicted vs.\ actual yield scatter under LOCO for the
           Prithvi-EO / Ridge condition (one panel per held-out country).
           The 1:1 line is shown dashed. Systematic under- and
           over-prediction reflects country-level yield distribution shift.}
  \label{fig:scatter}
\end{figure*}

Figure~\ref{fig:country_rmse} shows per-country LOCO RMSE across all conditions
and Figure~\ref{fig:scatter} plots predicted vs.\ actual yield for the Prithvi-EO / Ridge condition.
Rwanda is the hardest held-out country across all conditions
(RMSE $>$ 2{,}100 kg/ha), consistent with its unusually high
mean yield (3{,}993 kg/ha) relative to training-set means.
Nigeria similarly shows the largest naive-baseline deficit
(naive R$^2 = -2.121$), plausibly explained by the use of
admin-centroid proxy observations.
Kenya and Tanzania are easiest to predict (lowest RMSE under LOCO),
likely because their yield ranges overlap more substantially with
the pooled training distribution.

\subsection{Label Shift as Primary Bottleneck}

\begin{figure}[!t]
  \centering
  \includegraphics[width=\columnwidth]{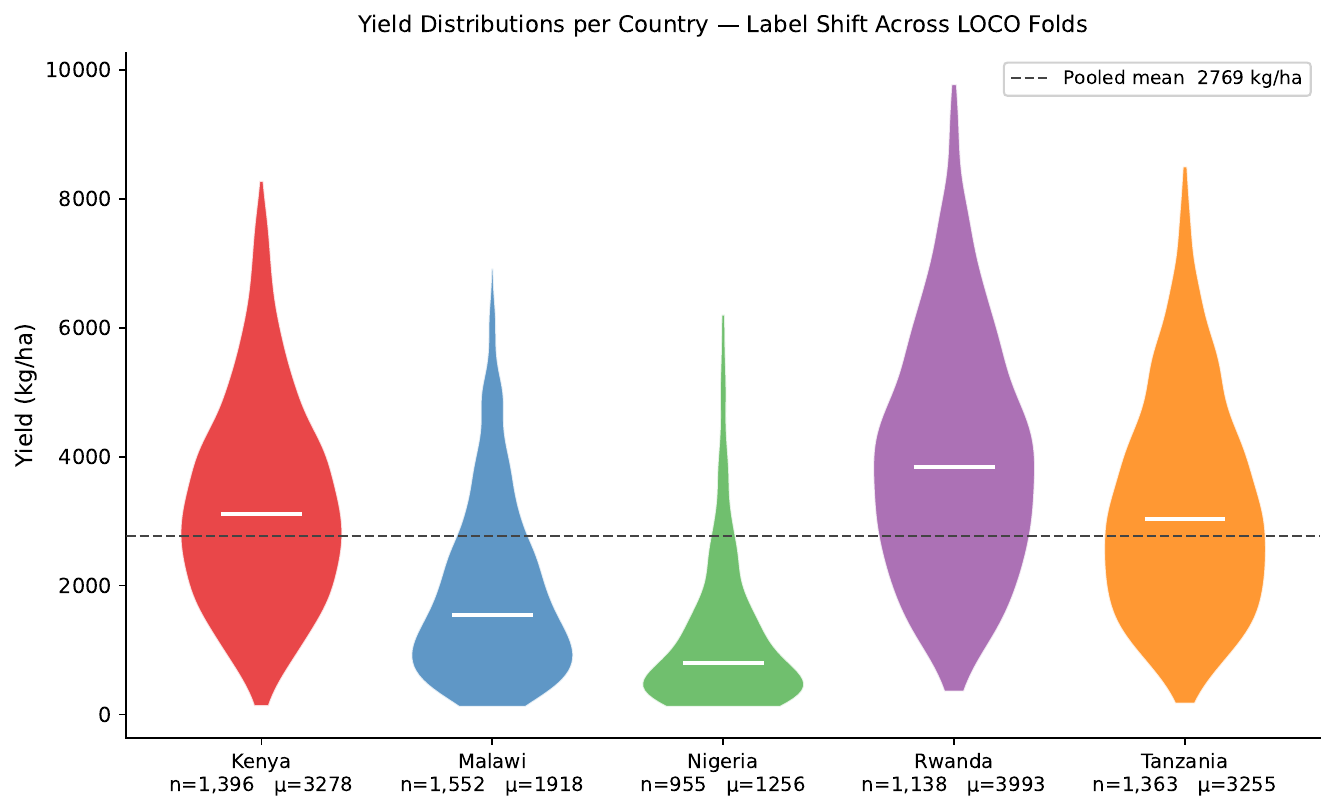}
  \caption{Yield distributions (kg/ha) per country. Nigeria and Rwanda exhibit
           markedly different central tendency and spread relative to the pooled
           training set, explaining why LOCO performance collapses for these folds
           regardless of feature representation.}
  \label{fig:yield_dist}
\end{figure}

\begin{figure}[!t]
  \centering
  \includegraphics[width=\columnwidth]{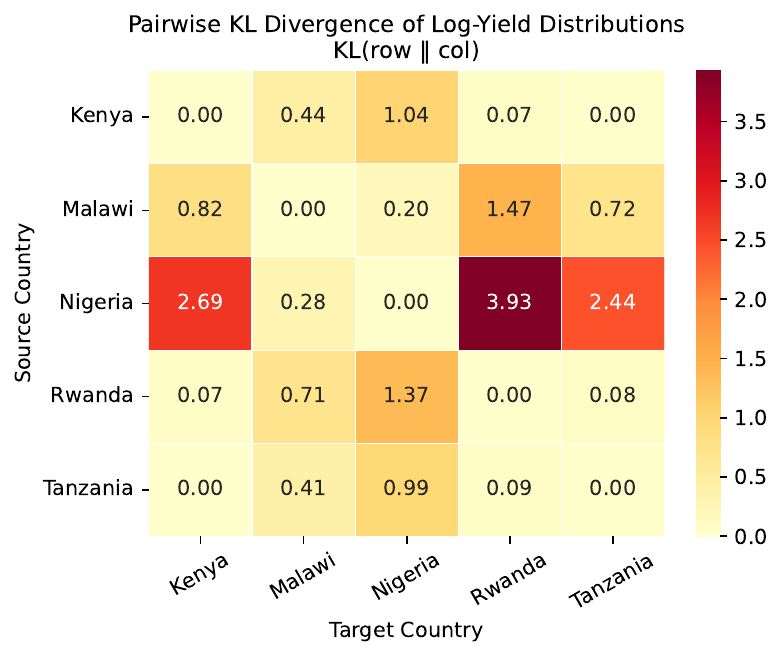}
  \caption{Pairwise KL divergence of log-yield distributions between countries
           (Gaussian approximation). Nigeria is the distributional outlier:
           KL(Nigeria\,$\Vert$\,Rwanda)\,$=3.93$, while Kenya--Tanzania divergence
           is near zero ($0.001$). Higher KL predicts worse LOCO performance.}
  \label{fig:kl}
\end{figure}

Figure~\ref{fig:yield_dist} visualises the yield distribution for each country.
Country means span from 1{,}256 kg/ha (Nigeria) to 3{,}993 kg/ha (Rwanda),
a factor of $3.2\times$ on the original scale.
Figure~\ref{fig:kl} quantifies pairwise label shift via Gaussian KL divergence
on log-yield distributions.
Nigeria is the distributional outlier:
KL(Nigeria $\|$ Kenya) $= 2.69$, KL(Nigeria $\|$ Rwanda) $= 3.93$.
Kenya and Tanzania are nearly identical: KL(Kenya $\|$ Tanzania) $= 0.001$.
The hardest held-out countries under LOCO (Nigeria, Rwanda) correspond exactly
to those with the largest KL divergence from the pooled training pool;
the easiest (Kenya, Tanzania) have near-zero divergence from each other.

\subsection{Statistical Separability of Conditions}

\begin{figure}[!t]
  \centering
  \includegraphics[width=\columnwidth]{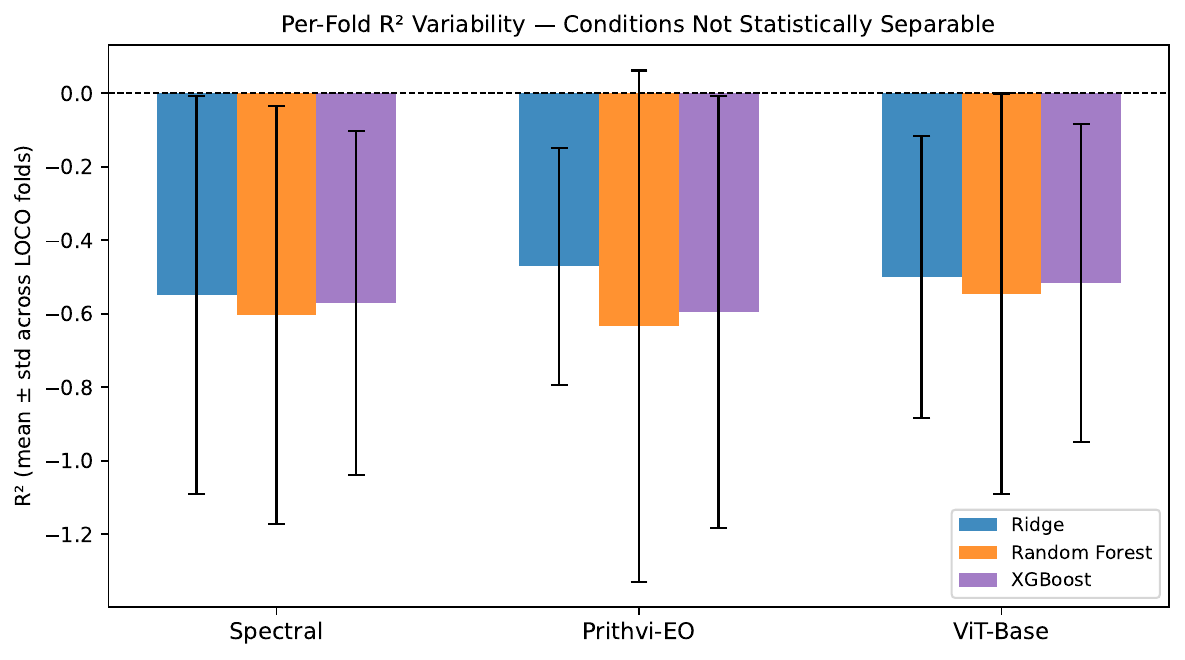}
  \caption{Mean LOCO R$^2$ $\pm$ one standard deviation across the five
           LOCO country folds, by feature set and regressor. Per-fold standard
           deviations (0.32--0.70) dwarf the between-condition differences
           ($<$0.07), indicating that no condition is statistically separable
           from any other.}
  \label{fig:errorbars}
\end{figure}

\begin{figure}[!t]
  \centering
  \includegraphics[width=\columnwidth]{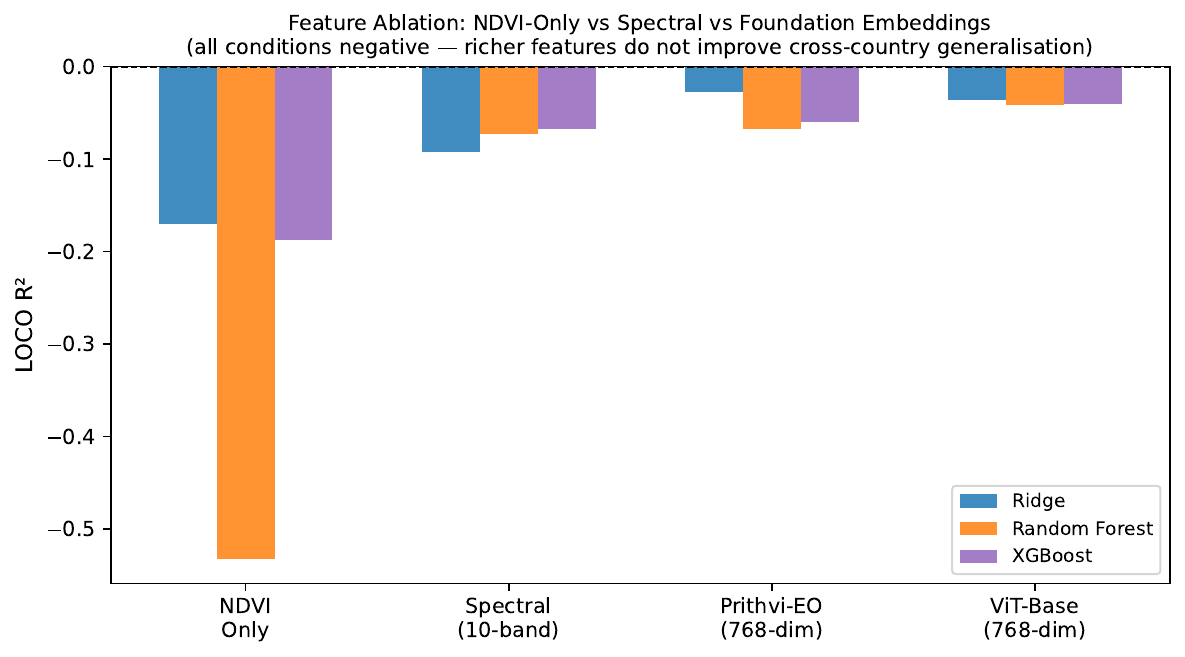}
  \caption{Feature ablation under LOCO: a single NDVI feature achieves R$^2$
           comparable to 768-dimensional Prithvi-EO or ViT-Base embeddings,
           confirming that richer representations do not improve cross-country
           generalisation under the frozen, single-frame protocol.}
  \label{fig:ablation}
\end{figure}

Figure~\ref{fig:errorbars} shows that the large within-fold variance renders
all nine conditions statistically indistinguishable.
Figure~\ref{fig:ablation} further shows that a single NDVI feature
(Ridge R$^2 = -0.170$) performs comparably to the full 768-dimensional
Prithvi-EO embeddings (Ridge R$^2 = -0.027$) — a gap well within
one per-fold standard deviation.
Together these results confirm that the bottleneck is distributional,
not representational.

\section{Discussion}
\label{sec:discussion}

\subsection{Why Foundation Models Do Not Close the Gap}

The core finding — that frozen Prithvi-EO embeddings applied to single
growing-season composites are not meaningfully superior to spectral features
under LOCO — requires precise qualification before drawing broader conclusions
about geospatial foundation models.

\textbf{Design mismatch.}
Prithvi-EO is a multi-temporal masked autoencoder pre-trained on six-channel
HLS time series. Its inductive biases — temporal self-attention, multi-frame
reconstruction — are optimised for time-series input.
By feeding a single annual composite we discard the phenological signal the
model was trained to exploit.
This is a deliberate experimental choice: it isolates the question of whether
frozen spatial representations alone carry cross-country invariance.
The answer is no, but this should not be interpreted as a general failure
of Prithvi-EO; multi-temporal fine-tuning may yield substantially different
results and remains an important open experiment.

\textbf{Geographic domain shift.}
Prithvi-EO was pre-trained predominantly on North American HLS tiles.
African smallholder landscapes — fragmented parcels, mixed cropping, and
informal field boundaries — differ substantially from the large-field
Corn Belt imagery that dominates the pre-training corpus.
This geographic domain gap adds a second source of distribution mismatch
beyond the temporal one.

\textbf{Label shift as the primary bottleneck.}
Section~\ref{sec:results} showed that the countries hardest to predict under
LOCO (Nigeria, Rwanda) are exactly those with the largest KL divergence from
the pooled training pool (Figure~\ref{fig:kl}), while the easiest (Kenya,
Tanzania) have near-zero mutual divergence.
This pattern holds regardless of feature set or regressor, and persists even
after removing Nigeria from both training and test folds
(Table~\ref{tab:nng}).
The conclusion is that no feature representation — however rich or
domain-specific — can recover accurate yield magnitudes for a held-out country
whose yield distribution occupies a fundamentally different range from the
training pool.
Addressing label shift requires country-level yield normalisation,
richer auxiliary covariates (soil fertility, management inputs, variety
adoption), or explicit distributional alignment — information not encoded
in satellite imagery alone.

\subsection{Implications for Benchmarking}

Our results demonstrate concretely that within-country random CV
inflates performance by 0.22--0.38 R$^2$ units relative to LOCO.
Practitioners and reviewers should treat any yield prediction accuracy
figures reported under random within-country CV with commensurate
scepticism when the intended application is cross-country or
cross-region deployment.
LOCO, or at minimum a spatial block CV scheme that prevents
geographic leakage, should be the default evaluation protocol
for operational yield prediction.

\subsection{Limitations}
\label{sec:limits}

Several limitations constrain the scope of our conclusions:

\begin{enumerate}
  \item \textbf{Frozen embeddings only.} Fine-tuning Prithvi-EO on African maize
        data may recover generalisation that frozen inference cannot.
        This remains an important open experiment.
  \item \textbf{Single-timestamp composites.} Prithvi-EO was designed for
        multi-temporal input; using a single growing-season composite
        discards phenological information the model was trained to exploit.
  \item \textbf{Nigeria label quality.} Admin-centroid proxy labels for Nigeria
        introduce spatial mismatch error relative to GPS-tagged fields.
  \item \textbf{Five-country sample.} LOCO over five countries produces five
        test folds. Per-country R$^2$ standard deviations of 0.32--0.70
        (Table~\ref{tab:foldstd}) far exceed the between-condition
        differences ($<0.07$); no formal significance test is applicable
        at this fold count, and all differences should be treated as
        directional only.
  \item \textbf{No domain adaptation.} Techniques such as domain-adversarial
        training, country-level normalisation, or meta-learning were not
        evaluated and may substantially improve cross-country transfer.
\end{enumerate}

\section{Conclusion}
\label{sec:conclusion}

We evaluated 18 combinations of feature representation, regression algorithm,
and cross-validation scheme for smallholder maize yield prediction across
five sub-Saharan African countries.
Within-country performance is moderate (R$^2$ up to 0.30) but
cross-country generalisation is uniformly poor (all LOCO R$^2 < 0$).
Frozen Prithvi-EO embeddings applied to single-frame composites provide no
meaningful advantage over 10-band Sentinel-2 spectral features.
This result should be understood as an evaluation of \emph{frozen, single-frame}
inference rather than a general indictment of geospatial foundation models:
multi-temporal fine-tuning on African data remains untested and is a promising
direction.
The evidence suggests that geospatial domain-specific pre-training alone —
without fine-tuning, multi-temporal input, or domain adaptation — is
insufficient to overcome country-level yield distribution shift.
Our work establishes a reproducible LOCO benchmark against which
future domain adaptation and meta-learning approaches for
African crop yield prediction should be compared.

\section*{Data and Code Availability}

The GROW-Africa yield labels are publicly available at
\href{https://doi.org/10.5281/zenodo.14961637}{doi:10.5281/zenodo.14961637}.
All preprocessing, embedding extraction, model training, and figure
generation code is available at
\url{https://github.com/yoadjei/yield-africa}.
Processed results (\texttt{results\_all.csv},
\texttt{results\_loco\_country.csv}) and paper figures are included in the
repository.

\section*{Acknowledgements}

The author thanks the GROW-Africa consortium for curating and releasing the
smallholder yield dataset, and IBM Research and NASA for releasing the
Prithvi-EO-1.0-100M model weights under an open licence.
Sentinel-2 imagery was accessed free of charge through the Google Earth Engine
research programme.

\textit{AI assistance disclosure:} Claude (Anthropic) was used as a coding
assistant to help develop and debug the Python pipeline scripts.
All experimental design, analysis, interpretation, and written text are
the author's own work.

\appendix
\section{Reproducibility Checklist}
\label{app:repro}

This appendix provides the information required to exactly reproduce all
reported numbers.

\subsection{Software Versions}

\begin{itemize}
  \item Python 3.10
  \item NumPy 1.26, pandas 2.2, scikit-learn 1.4, XGBoost 2.0
  \item PyTorch 2.2 (embedding extraction only), timm 0.9
  \item earthengine-api 0.1 (GEE patch export only)
\end{itemize}

\subsection{Random Seeds}

All stochastic operations use \texttt{random\_state=42}:
Random Forest (\texttt{n\_estimators=100}),
XGBoost (\texttt{random\_state=42}),
KFold shuffling for random CV (\texttt{KFold(shuffle=True, random\_state=42)}).
Ridge regression (RidgeCV) is deterministic given fixed features.

\subsection{Train/Test Split Logic}

\textbf{LOCO:} For each held-out country $c$, the training set is all observations
with \texttt{country} $\neq c$; the test set is all observations with
\texttt{country} $= c$.
No shuffling. No stratification. Splits are fully determined by the
\texttt{country} column in \texttt{master\_dataset.parquet}.

\textbf{Random CV:} Scikit-learn \texttt{KFold(n\_splits=5, shuffle=True, random\_state=42)}
applied to the full pooled dataset after dropping NaN yield rows.

\subsection{Preprocessing}

NaN features are imputed with the training-set column median
(computed after country split for LOCO; computed per fold for random CV).
Columns that are all-NaN in the training set are set to 0.
The yield target is log-transformed (\texttt{np.log1p}) prior to model fitting;
all reported RMSE and MAE values are back-transformed via \texttt{np.expm1}.

\subsection{Reproducing Each Table}

\begin{itemize}
  \item Tables~\ref{tab:random} and~\ref{tab:loco}: \texttt{python scripts/04\_train\_eval.py}
        $\rightarrow$ \texttt{data/processed/results\_all.csv}
  \item Table~\ref{tab:naive}: \texttt{python scripts/05\_figures.py} (naive baseline
        is computed inline during figure generation)
  \item Tables~\ref{tab:ndvi}, \ref{tab:nng}, and~\ref{tab:foldstd}:
        \texttt{python scripts/04b\_sensitivity.py}
        $\rightarrow$ \texttt{data/processed/results\_ndvi\_only.csv},
        \texttt{data/processed/results\_loco\_no\_nigeria.csv},
        \texttt{data/processed/results\_loco\_fold\_std.csv}
\end{itemize}

\subsection{Compute Requirements}

\begin{itemize}
  \item Steps 1--3 (data download, GEE export, CHIRPS): network-bound, \textless 2 hours
  \item Step 4 (preprocessing): CPU, \textless 5 minutes
  \item Step 5 (embedding extraction): GPU recommended (NVIDIA A100: $\approx$2~h;
        CPU: $\approx$8~h)
  \item Step 6 (train and evaluate, all conditions): CPU, $\approx$20 minutes
  \item Step 7 (sensitivity analyses): CPU, $\approx$25 minutes
\end{itemize}

\bibliographystyle{IEEEtran}
\bibliography{references}

\end{document}